\documentclass[10pt,twocolumn,letterpaper]{article}

\usepackage{booktabs}
\usepackage{graphicx}
\usepackage{array}
\usepackage{multirow}
\usepackage{makecell}
\usepackage[table]{xcolor}

\usepackage{pifont}
\newcommand{\cmark}{\ding{51}}
\newcommand{\xmark}{\ding{55}}

\newcolumntype{L}{>{\raggedright\arraybackslash}m{2cm}}
\newcolumntype{C}{>{\centering\arraybackslash}m{1.2cm}}
\newcolumntype{G}{>{\centering\arraybackslash}m{1.8cm}}

\newcolumntype{X}{>{\centering\arraybackslash}m{0.6cm}}
\newcolumntype{Y}{>{\centering\arraybackslash}m{0.8cm}}
\newcolumntype{Z}{>{\centering\arraybackslash}m{1.5cm}}
\usepackage[pagenumbers]{cvpr} 

\newcommand\mypara[1]{\noindent\textbf{#1.}}

\definecolor{cvprblue}{rgb}{0.21,0.49,0.74}
\usepackage[pagebackref,breaklinks,colorlinks,allcolors=cvprblue]{hyperref}

\def\paperID{} 
\def\confName{}
\def\confYear{}

\title{SecAgent: Efficient Mobile GUI Agent with \textit{Se}mantic \textit{C}ontext}

\author{
Yiping Xie\thanks{Equal contribution.},
Song Chen$^*$,
Jingxuan Xing$^*$,
Wei Jiang$^*$,
Zekun Zhu$^*$, \\
Yingyao Wang, 
Pi Bu, 
Jun Song, 
Yuning Jiang\thanks{Corresponding author.},
Bo Zheng \\
Taobao \& Tmall Group of Alibaba \\
{\small \{yizhou.xyp, jingxuan.xjx, zekun.zhu, mengzhu.jyn\}@alibaba-inc.com}}

\begin{document}
\maketitle
\begin{abstract}
Mobile Graphical User Interface (GUI) agents powered by multimodal large language models have demonstrated promising capabilities in automating complex smartphone tasks.
However, existing approaches face two critical limitations: the scarcity of high-quality multilingual datasets, particularly for non-English ecosystems, and inefficient history representation methods.
To address these challenges, we present SecAgent, an efficient mobile GUI agent at 3B scale.
We first construct a human-verified Chinese mobile GUI dataset with 18k grounding samples and 121k navigation steps across 44 applications, along with a Chinese navigation benchmark featuring multi-choice action annotations.
Building upon this dataset, we propose a semantic context mechanism that distills history screenshots and actions into concise, natural language summaries, significantly reducing computational costs while preserving task-relevant information.
Through supervised and reinforcement fine-tuning, SecAgent outperforms similar-scale baselines and achieves performance comparable to 7B-8B models on our and public navigation benchmarks.
Our dataset is available at \url{https://huggingface.co/datasets/alibabagroup/CMGUI}.
\end{abstract}
\section{Introduction}
\label{sec:intro}

\begin{figure}[!t]
\centering
\includegraphics[width=0.95\linewidth]{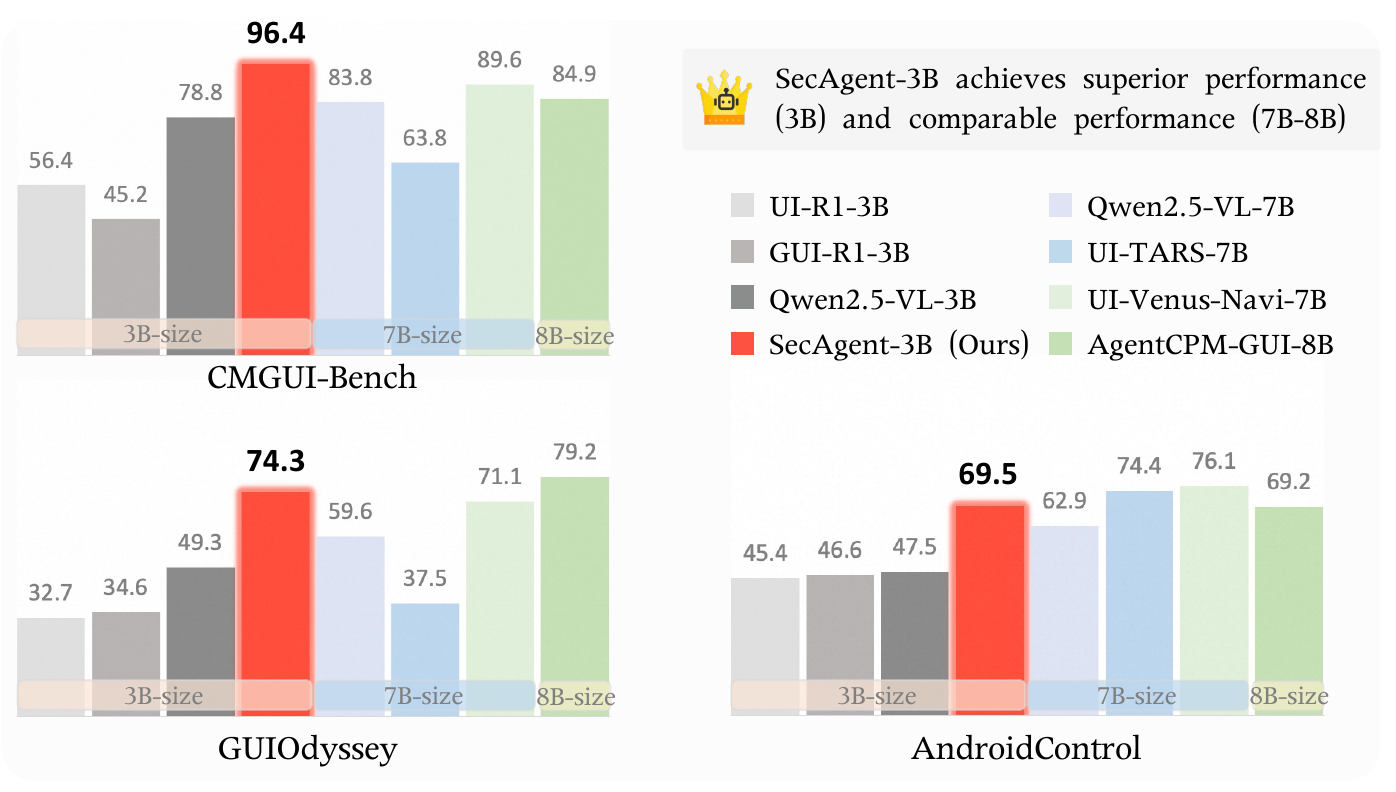}
\vspace{-6pt}
\caption{Performance comparison of various agent models on three representative navigation benchmarks. SecAgent-3B achieves superior performance to 3B models and comparable performance to 7B-8B models.
}
\label{fig:compare_performance}
\vspace{-16pt}
\end{figure}

\begin{table*}[htbp]
\centering
\footnotesize
\begin{tabular}{l XXXXYY l XXXX}
\toprule
\multirow{2}{*}{\textbf{Name}} & \multicolumn{6}{c}{\textbf{Training Data}} & \multirow{2}{*}{\textbf{Name}} & \multicolumn{4}{c}{\textbf{Benchmark Data}} \\
\cmidrule(lr){2-7} \cmidrule(lr){9-12}
 & Traj. & Steps & Apps & CoT & Human-Verified& Open-Source & &
 Traj. & Steps & Apps & Multi-Choice \\
\midrule
Mobile-Bench-v2 \cite{xu2025mobile} & 12,856 & 93,592 & 49 & \xmark & \xmark & \cmark & Magic-RICH \cite{tang2025magicguifoundationalmobilegui} & - & 4,000 & 150 & \xmark \\
AgentCPM-GUI \cite{zhang2025agentcpm}& 55k & 470k & 30 & \xmark & \cmark & \xmark & CAGUI \cite{zhang2025agentcpm} & 600 & 4,516 & 30 & \xmark \\

CMGUI (Ours) & 29,711 & 121,265 & 44 & \cmark & \cmark & \cmark & CMGUI-Bench (Ours) & 390 & 2,574 & 44 & \cmark \\
\bottomrule
\end{tabular}
\vspace{-6pt}
\caption{Comparison between our CMGUI dataset and existing Chinese mobile navigation datasets and benchmarks. We consider the scale (the number of trajectories and step annotations), diversity (the number of apps), whether each step includes a thought annotation, whether the dataset is verified by humans, whether it has been open-sourced, and whether the benchmark has multi-choice annotations. Note that Magic-RICH only contains step-level data.} 
\label{tab:gui_benchmarks_zh}
\vspace{-16pt}
\end{table*}

Mobile Graphical User Interface (GUI) agents powered by Large Language Models (LLMs) and Multimodal LLMs (MLLMs) have significantly evolved in recent years \cite{tang2025survey, liu2025llm}. 
Early text-based approaches \cite{wang2023enabling, wen2024autodroid, wen2023droidbot, lee2023explore} rely on structured metadata such as accessibility trees, but face limitations, including dependency on platform-specific parsing and vulnerability to incomplete metadata. 
Recognizing these constraints, the community has shifted toward vision-based approaches that perceive interfaces directly through screenshots, mirroring human visual interaction 
\cite{lin2025showui, cheng2024seeclick, huang2025spiritsight}.
These vision-based agents exhibit cross-platform compatibility and can handle diverse UI designs without requiring structured metadata, demonstrating strong capabilities in element grounding and multi-step navigation.

Despite these achievements, current mobile GUI agent research faces two critical limitations.
First, existing work predominantly focuses on the English mobile ecosystem \cite{wu2024atlas, hong2024cogagent}, with limited attention devoted to non-English environments. 
Building robust multilingual GUI agents fundamentally requires diverse multilingual and cross-cultural data with ecosystem-specific knowledge, such as element functionality and interaction conventions. 
However, the scarcity of open-source non-English datasets poses a substantial barrier. 
While several recent studies \cite{zhang2025agentcpm, tang2025magicguifoundationalmobilegui, zeng2025uitron} have constructed large-scale Chinese mobile datasets, these datasets remain proprietary and inaccessible to the research community.
To advance research on multilingual GUI agents, we introduce \textbf{CMGUI}, a large-scale Chinese mobile GUI dataset, which comprises 18k grounding samples and 29k navigation episodes with 121k navigation steps across 44 diverse applications. 
To ensure the data quality, the navigation step undergoes rigorous human verification with manual bounding box annotations. 
Building upon CMGUI, we further introduce \textbf{CMGUI-Bench}, a high-quality Chinese mobile navigation benchmark. 
To accommodate diverse GUI manipulations, each step in CMGUI-Bench is annotated with multiple valid choices (\ie, actions) when available.
Tab. \ref{fig:compare_performance} presents a comparison with existing Chinese datasets and benchmarks. 

Second, existing methods lack efficient history representation for multi-step navigation \cite{lu2024gui,qin2025ui, wang2025ui}. 
Informed actions at each step typically rely on historical information such as prior screenshots and actions.
However, complete history, particularly screenshots, causes prohibitive computational overhead, with each screenshot being encoded into thousands of visual tokens \cite{hong2024cogagent}. 
Consequently, current approaches adopt simplified history representations: some methods \cite{chen2024guicourse, wang2025ui, lin2025showui} restrict the history window to recent steps (\eg, the last 5 steps), while others \cite{wu2024atlas, xuaguvis, gu2025ui, luo2025gui} retain only action sequences. 
The former approach inevitably suffers from information loss, especially when the history window is small; the latter provides insufficient information, as parameterized actions (\eg, click coordinates) without their associated screenshots cannot convey meaningful historical progression.
To address this challenge, we propose the \textbf{Semantic Context Mechanism}, which presents the historical information as a semantic context, a concise natural language-based summary that records critical information from previous interactions.
By dynamically maintaining the semantic context across navigation steps, the agent avoids accessing the entire history screenshots and actions, significantly improving the computational efficiency while preserving completeness of the historical information.
Furthermore, we present \textbf{SecAgent} 
based on the semantic context mechanism.
Through supervised and reinforcement fine-tuning, SecAgent achieves highly competitive performance as shown in Fig. \ref{fig:compare_performance}.
It outperforms prior methods on CMGUI-Bench and matches 7B-8B models on AndroidControl \cite{li2024effects} and GUIOdyssey \cite{lu2024gui}.

Our contributions are summarized as follows:
\begin{itemize}
    \item We introduce CMGUI, a large-scale high-quality Chinese mobile GUI dataset, and CMGUI-Bench, a navigation benchmark with multi-choice annotations (Sec. \ref{sec:dataset}).
    \item We propose the semantic context mechanism, which efficiently represents critical historical information (Sec. \ref{sec:method}).
    \item Extensive experiments demonstrate the superiority of SecAgent in both our and public benchmarks (Sec. \ref{sec:exp}).
\end{itemize}
\section{Related Works}
\label{sec:relate_works}

\begin{figure*}[!htbp]
\centering
\includegraphics[width=0.92\linewidth]{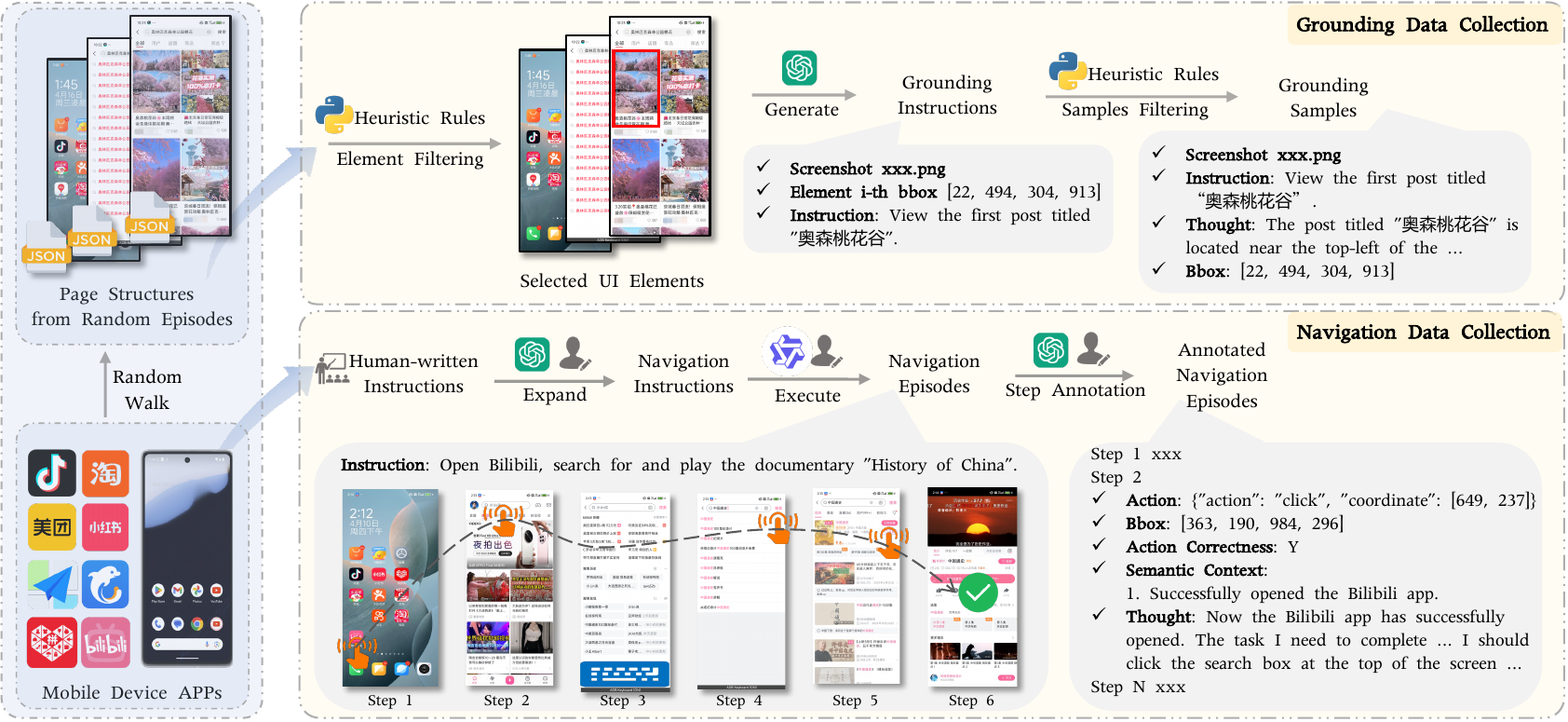}
\caption{
Collection pipeline of our CMGUI dataset. Grounding data: UI elements are selected from random walk episodes, with MLLMs generating instructions. Navigation data: episodes are collected using a hybrid human-agent strategy, then humans annotate action correctness and bounding boxes, while MLLMs annotate semantic context and thoughts.
}
\label{fig:dataset}
\vspace{-14pt}
\end{figure*}

\mypara{GUI Benchmarks and Datasets}
Numerous benchmarks and datasets have been introduced to advance research on mobile GUI navigation.
Some works \cite{xu2024androidlab, rawles2024androidworld, xing2024understanding, 
toyama2021androidenv, zhang2023mobile, zhang2025appagent, chai2025a3, zhang2025agentcpm, tang2025magicguifoundationalmobilegui} design benchmarks to comprehensively evaluate agents’ navigation capabilities, while others \cite{li2020mapping, burns2021mobile, sun2022meta, venkatesh2022ugif, rawles2023androidinthewild, zhang2024android, li2024effects, chai2024amex, lu2024gui, xuaguvis, liu2025learnact, zhang2025tongui, xu2025mobile}
curate datasets to enhance the perception, grounding, planning, and reasoning capabilities.
However, existing studies primarily focus on the English mobile ecosystem, and limited studies \cite{xu2025mobile, tang2025magicguifoundationalmobilegui, zhang2025agentcpm} focus on the Chinese ecosystem.
The comparison of our proposed CMGUI and CMGUI-Bench with existing Chinese mobile navigation datasets and benchmarks is summarized in Tab. \ref{tab:gui_benchmarks_zh}.
Notably, CMGUI features human verification to ensure data quality, and CMGUI-Bench with multi-action annotations accommodates diverse GUI manipulations.

\mypara{GUI Agents}
LLM-powered GUI agents generally fall into two paradigms: prompting-based and training-based.
Prompting-based agents directly query existing general models \cite{achiam2023gpt, Anthropic_Claude3_2024} to generate GUI actions with carefully constructed prompts that contain the instruction, GUI state, and action space.
For example, some works \cite{wang2023enabling, wen2024autodroid, wen2023droidbot, lee2023explore} query LLMs with textual prompts, while others  \cite{yan2023gpt, zhang2025appagent, lu2024omniparser} incorporates visual inputs and query MLLMs.
Training-based approaches fine-tune
 open-source foundation models \cite{bai2025qwen2,chen2024internvl, liu2023visual} to obtain more proficient GUI agents. 
Through SFT on large-scale and multi-level datasets, the fine-tuned agents \cite{lin2025showui, cheng2024seeclick, wu2024atlas, huang2025spiritsight} demonstrate superior grounding and navigation capabilities across diverse platforms.
More recently, RFT, combining rule-based rewards with policy updates (\eg, GRPO \cite{shao2024deepseekmath}), is further incorporated to enhance agents’ reasoning and generalization capabilities \cite{liu2025infigui, zhou2025gui, lu2025ui, gu2025mobile}.
Our work also employs a two-stage paradigm, incorporating SFT and RFT, but proposes a novel history representation mechanism that differs from existing methods.

\section{CMGUI Dataset}
\label{sec:dataset}
In this section, we introduce the constructed CMGUI and the corresponding benchmark.
An overview of the entire collection process is illustrated in Fig. \ref{fig:dataset}. 
\footnote{In this and subsequent figures, Chinese text has been translated into English for presentation purposes.}

\subsection{Grounding Data Collection} \label{subsec:grounding}

The grounding data is designed to equip agents with the ability to localize and understand specific UI elements based on functional descriptions \cite{zhang2025agentcpm, wang2025ui, qin2025ui}.
The collection pipeline contains the following three stages:

\mypara{Raw Data Acquisition} 
To populate the grounding data at scale, we perform the random walk strategy \cite{wu2024mobilevlm} to explore various system surfaces and apps.
During this autonomous exploration, the executed action and the device state, comprising the screenshot and the page structure data, are captured at every step. 
Here, the page structure data provides a structured, hierarchical representation of the UI elements on the mobile screen, including bounding boxes and descriptive attributes for all potentially interactive UI elements.

\mypara{UI Element Filtering} 
To ensure unambiguous and intent-aligned instructions, we implemented a rigorous pipeline for filtering UI elements. Elements are filtered based on geometric (\eg, area $<$ 6,000 pixels, aspect ratio $>$ 13.5, or occupying $>$15\% of screen) and visual (near-uniform color) heuristics. Only leaf nodes in the page hierarchy are retained to enforce semantic atomicity. Spatial redundancy is controlled via reservoir sampling, where previously seen elements are downsampled to a 5\% selection probability.

\mypara{Instruction Generation} 
For each retained element, we overlay a visible marker on the screenshot and use GPT-4 to generate an imperative and unambiguous grounding instruction, along with the rationale of the instruction.
Then, a quality check is performed to filter out samples where the generated instructions are shorter than four words or their rationale is fewer than ten words.
Finally, we obtain 18,887 high-quality grounding samples, encompassing eight widely used applications such as Taobao and Rednote.

\subsection{Navigation Data Collection} \label{subsec:navigation}
The navigation data focuses on enhancing the multi-step planning ability of GUI agents, which is constructed using a hybrid human-agent collaboration strategy.

\mypara{Action Space}
Our action space contains $\texttt{CLICK}$, $\texttt{SWIPE}$, $\texttt{TYPE}$, $\texttt{SYSTEM\_BUTTON}$,
$\texttt{WAIT}$,
and $\texttt{TERMINATE}$.
Details are provided in the supplementary material.

\mypara{Instruction Generation}
We construct high-level instructions through a hybrid human-LLM workflow. First, we prompt GPT-4 with app-specific functional descriptions to generate initial instruction candidates. 
Then, these candidates are rigorously reviewed and refined by human annotators to ensure feasibility, clarity, and realism. 
To increase linguistic diversity, we further apply paraphrasing and contextual expansion using GPT-4, resulting in a rich and varied instruction pool.

\mypara{Hybrid Instruction Execution} 
To reduce the cost of human demonstration, we employ a hybrid human-agent collaboration strategy to execute instructions.
First, a subset of instructions is executed manually by annotators to generate successful navigation episodes with high-quality ground-truth.
Subsequently, the annotated episodes are used to train an auxiliary agent by fine-tuning a compact MLLM (Qwen-2.5-VL-3B).
It is then deployed to scale data collection through agent-driven execution of the remaining instructions, limited by a maximum of 30 steps each instruction.
Here, the auxiliary agent executes each instruction multiple times to increase the diversity of collected episodes.
\begin{figure}[thbp]
\centering
\includegraphics[width=0.95\linewidth]{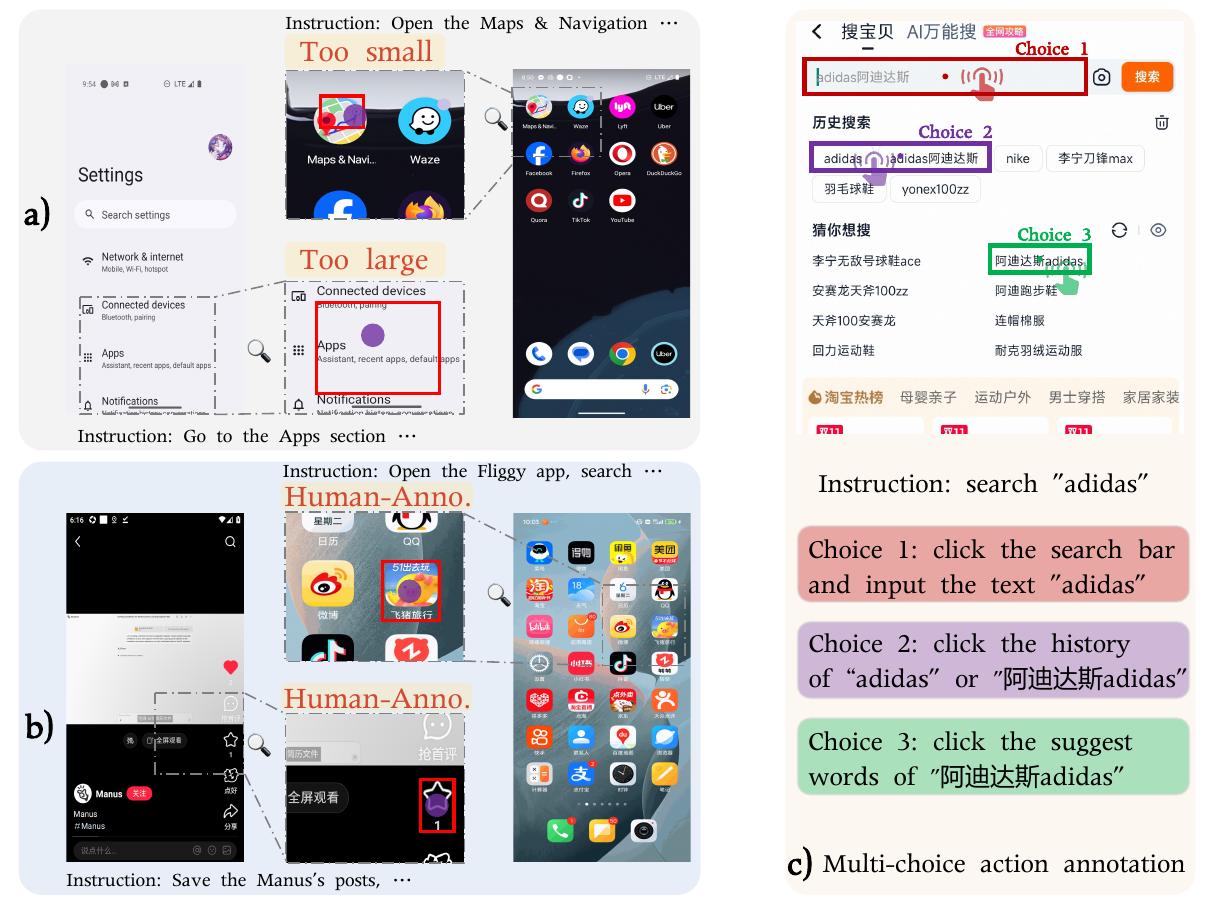}
\caption{a) Samples from the GUIOdyssey dataset \cite{lu2024gui}, where SAM2 \cite{ravi2024sam} is used to segment UI elements to obtain bounding boxes, inevitably leading to noise and errors (\eg, the red box is incorrectly sized). b) Samples from our CMGUI data. We employ well-trained annotators to conduct box-by-box annotation and implement a double-check mechanism to ensure precise bounding boxes. c) For our CMGUI-Bench, we conduct step-by-step and box-by-box annotation and consider all reasonable actions at each step, producing multi-choice action annotation.}
\label{fig:dataset_samples}
\vspace{-14pt}
\end{figure}
\mypara{Episode Annotation}
We collect episodes executed by both humans and agents, followed by a step-by-step annotation process. Starting from the first step of each episode, annotators verify the correctness of actions until either termination or the first error is encountered.
For each confirmed correct click action, annotators provide a precise bounding box. At the first incorrect step, the correct action and its bounding box are annotated.
Notably, our human-annotated bounding boxes demonstrate higher precision compared to those generated by autonomous approaches (\eg, SAM2 \cite{ravi2024sam} in GUIOdyssey \cite{lu2024gui}).
Representative examples are illustrated in Fig. \ref{fig:dataset_samples} (a)-(b).
After human verification and correction, we use GPT-4o \cite{hurst2024gpt} to augment the annotated steps with two additional fields: (1) semantic context, summarizing critical information from previous steps, and (2) thought, describing the reasoning behind the action.
Furthermore, for instructions with a low completion rate, a fallback human execution phase is employed.

To scale data collection efficiently, we adopted a \textit{data flywheel} strategy: in each iteration, after collecting and annotating a new batch of trajectories, we retrain the auxiliary agent,  leveraging all previously accumulated data. The detailed training methodology is presented in Sec. \ref{sec:method}.
In total, we collect 29,711 trajectories comprising 192,666 steps, of which 121,265 steps are annotated. 
We train exclusively on annotated data; in particular, any step data following the first erroneous action is discarded from the training corpus.
The training data covers 44 unique mobile applications across key functional domains such as e-commerce (\eg, Pinduoduo, JD), social media (\eg, Zhihu, Bilibili), and local services (\eg, Meituan, Ele.me).

\begin{figure*}[thbp]
\centering
\includegraphics[width=0.98\linewidth]{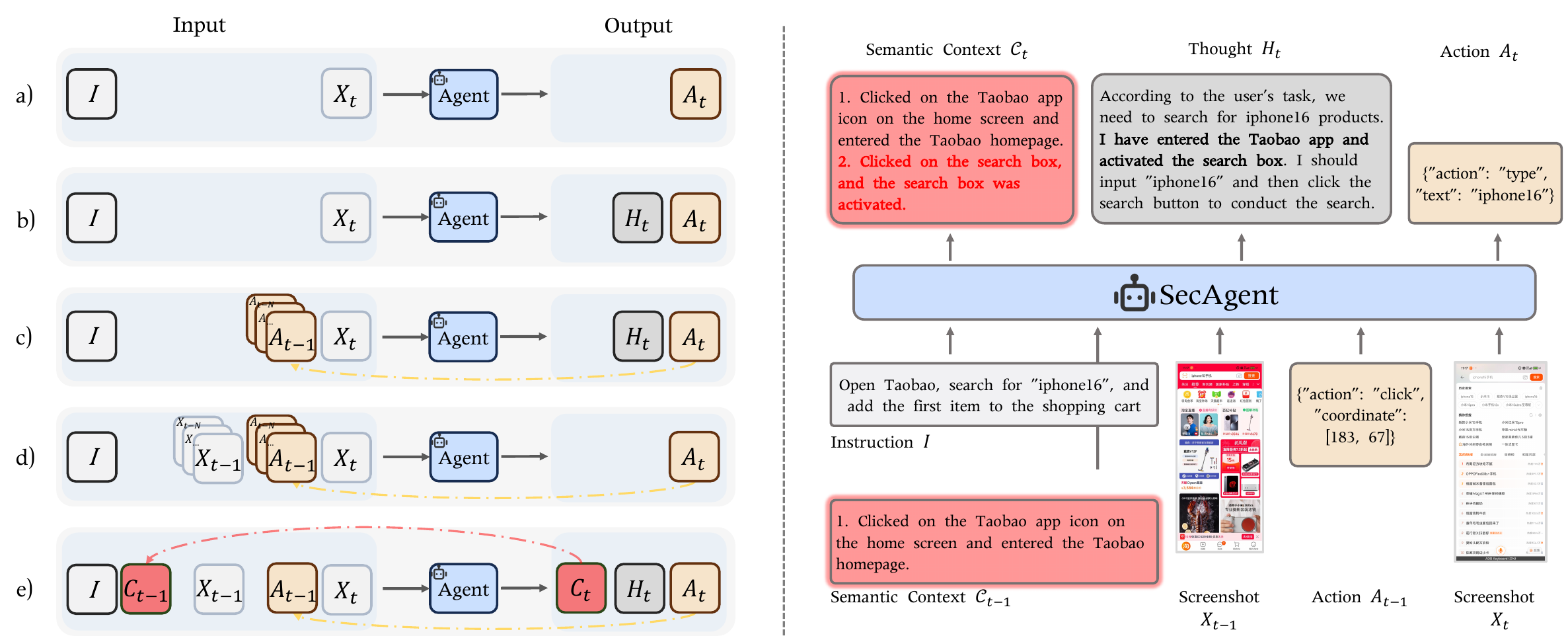}
\vspace{-4pt}
\caption{Comparison of different methods. \textbf{Left.} a) Baseline: take the instruction and the current screenshot as input and directly output the action. b) AgentCPM \cite{zhang2025agentcpm} and UI-R1 \cite{lu2025ui} expect the model to have reasoning capabilities and thus introduce thought before output action. c) UI-Venus \cite{qin2025ui}, GUI-R1 \cite{luo2025gui}, and OS-Atlas \cite{wu2024atlas} add history actions to the input. d) OdysseyAgent \cite{lu2024gui} uses both history actions and screenshots in the input. To ensure optimal results, it uses a large number of historical screenshots, for example, $N=5$. e) Our SecAgent records history information by maintaining a concise and natural language-based summary. Thanks to this, SecAgent can achieve good results using only a historical image and action. \textbf{Right.} A demonstration of SecAgent.}
\label{fig:model}
\vspace{-12pt}
\end{figure*}

\subsection{CMGUI Benchmark}
To evaluate the agent's navigation capability in Chinese mobile applications, we introduce CMGUI-Bench from our curated navigation corpus. The benchmark includes only successful episodes with reliable ground truth. To ensure diversity, we filter episodes using a programming-distance criterion: each instruction must be at least six units distant from others, promoting broad coverage of user intents and interaction patterns while avoiding near-duplicates.

CMGUI-Bench comprises 390 complete episodes and 2,574 steps, with an average episode length of 6.6, spanning 44 widely used apps in the Chinese ecosystem. During curation, we observe that many steps admit multiple valid actions due to interface affordances and app-specific flows \cite{chen2025ui}. Typical cases include submitting a query by either tapping a search button or selecting a suggestion, and reaching a target page via either clicking a tag or swiping horizontally. Some examples are shown in Fig. \ref{fig:dataset_samples} (c). We therefore recheck all steps and annotate alternative actions to accommodate diverse GUI manipulations. Evaluation focuses on four actions: click, type, swipe, and terminate. A click is correct if it lands within the target bounding box, typed content must match the ground truth, and swipes must align in the direction of the annotated intent. 

\section{Method: SecAgent}
\label{sec:method}

\subsection{Formulation}
We adopt an agent-based perspective \cite{qin2025ui, wang2025ui}, where an agent is formalized as a parameterized policy that maps instructions, context, and environment state to actions. At each timestep $t$, the agent follows the ReAct paradigm \cite{yao2022react}, iteratively reasoning, acting, and processing feedback.

\mypara{Definitions}
Instruction ($I$) is the natural language task query prompting the agent. Action ($A_t$) is the agent's behavior at time $t$ (\eg, GUI manipulation, system commands). Screenshot ($X_t$) is the environment state at time $t$. Thought ($H_t$) is the agent's internal reasoning and planning. Semantic context ($C_t$) is a natural language summary of key prior operations, detailed in Sec. \ref{subsec:architecture}.

A \textit{grounding sample} is defined as $(I_{G}, X_{G}, A_{G})$, where $I_{G}$ is a low-level instruction (\eg, ``click the search button''), $X_{G}$ is the screenshot, and $A_{G}$ is the bounding box of the target area.
A \textit{navigation episode} is a sequence that completes a high-level instruction: $E = (I_{N}, \{(X_t, A_t)\}_{t=1}^{T})$, where $I_{N}$ is the instruction (\eg, ``set an alarm for 8 am tomorrow'') and $T$ is the number of steps in the episode.
The baseline agent \cite{hong2024cogagent} can be formalized as:
\begin{equation}
P(A_t \mid I, X_t)
\end{equation}
Furthermore, we can use historical screenshots and actions \cite{lu2024gui, wang2025ui} as follows:
\begin{equation}\label{eq2}
P(A_t \mid I, X_t, \{X_{t-i}, A_{t-i}\}_{i=1}^{N})
\end{equation}
where $N$ denotes the size of the history window.
Additionally, we can incorporate thoughts to enhance the agent's reasoning capability \cite{zhang2025agentcpm, lu2025ui}.
The comparison of existing methods is illustrated in Fig. \ref{fig:model} (Left).

\subsection{Architecture} \label{subsec:architecture}
The completion of GUI navigation tasks requires a precise sequence of multi-step actions, whose correctness critically depends on the \textit{historical information} (\eg, previous screenshots and actions).
For instance, in GUI search tasks, clicking a search result depends on both the prior query and the history of visited pages and clicked links.
However, incorporating all raw historical information incurs an efficiency challenge \cite{jiang2025lightagent}. An action description requires only a handful of tokens, whereas an MLLM-embedded screenshot occupies hundreds of tokens. Stacking all the raw historical information as shown in Eq. (\ref{eq2}) leads to substantial \textit{computational overhead}, resulting in increased training costs and detrimental real-time inference latency.

To reduce the computational overhead caused by historical information while preserving the essential context required for task execution, we propose the \textbf{Semantic Context Mechanism}. 
A comparison of our proposed mechanism with existing history representation methods is illustrated in Fig. \ref{fig:model} (Left).   
At each step $t$, we augment the input with a textual context $C_t$ that summarizes critical information from previous steps, and a thought $H_t$ that provides the rationale for adopting action $A_t$.
The semantic context is maintained by the agent as a concise, natural language-based summary that records key operations from previous steps (\eg, clicked labels, typed inputs, searched outcomes).
This textual representation efficiently carries the essential history without requiring the model to process all past screenshots.
Due to this delicate design, we find that we can achieve good performance only using the last screenshot and action as historical information.
That is, $N=1$ in Eq. (\ref{eq2}).
Additionally, we also adopt the CoT mechanism to enhance the agent's reasoning and planning capabilities.
Formally, our proposed SecAgent predicts the next GUI manipulation as:
\begin{equation}\label{eq3}
P(C_{t}, H_t, A_t \mid I, X_t, C_{t-1}, X_{t-1}, A_{t-1})
\end{equation}
Specifically, SecAgent receives instruction $I$, current screenshot $X_t$, previous semantic context $C_{t-1}$, screenshot $X_{t-1}$, and action $A_{t-1}$. It then autoregressively generates the updated semantic context $C_{t}$ capturing critical events from the last interaction, $H_t$ providing reasoning and planning, and $A_t$ representing the grounded GUI manipulation.
An illustrative example is presented in Fig. \ref{fig:model} (Right).

\begin{figure*}[tbp]
\centering
\includegraphics[width=0.95\linewidth]{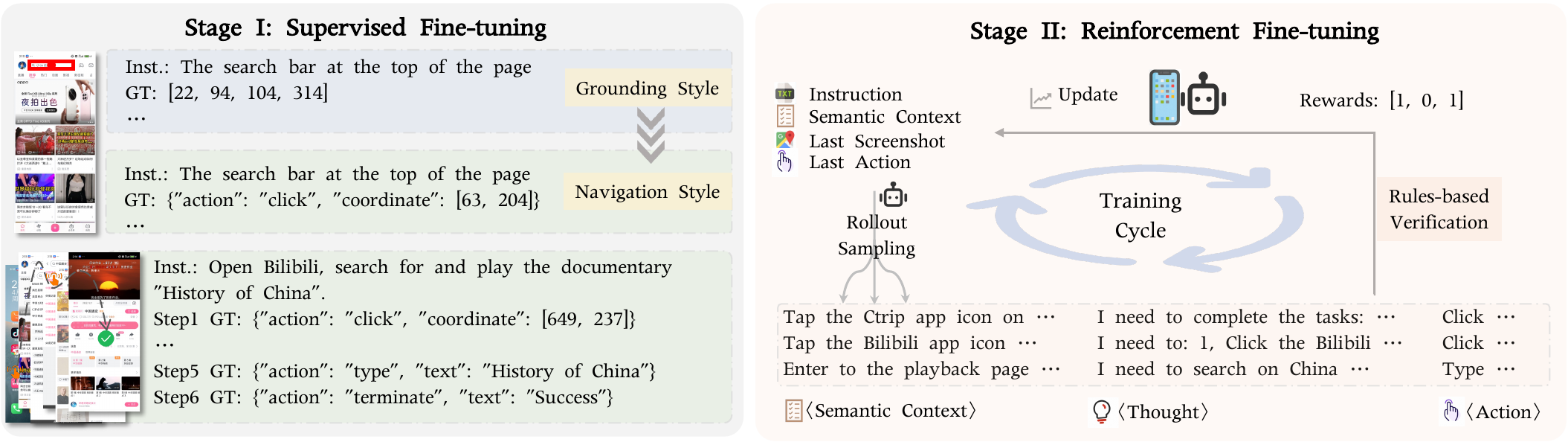}
\caption{The training framework of SecAgent. We observe that the original grounding and navigation datasets exhibit inherent biases. To mitigate this issue, we do not directly utilize the grounding data for SFT training as in prior work \cite{zhang2025agentcpm, zhang2025tongui}. Instead, we transform the grounding annotations into navigation-compatible format by treating the center point of each bounding box as the target click coordinate.}
\label{fig:three_stages}
\vspace{-5pt}
\end{figure*}
\begin{table*}[tbp]
\centering
\footnotesize
\begin{tabular}{lcccccc}
\toprule
\multirow{2}{*}{\textbf{Model}} & 
\multirow{2}{*}{\textbf{Step Acc.}} & 
\multirow{2}{*}{\textbf{Task Acc.}} &
\multicolumn{4}{c}{\textbf{Action Acc.}} \\
\cmidrule(lr){4-7}
 & & & Click Acc. & Type Acc. & Swipe Acc. & Terminate Acc. \\
\midrule
\multicolumn{4}{l}{\textit{\textbf{Closed-source Models}}} \\
GPT-4o \cite{hurst2024gpt}   & 48.4 & 0.0 & 43.5 & 68.3 & 63.7 & 62.1 \\
Claude-3-Sonnet \cite{Anthropic_Claude3_2024} & 81.4 & 25.8 & 91.1 & 30.0 & 81.3 & 53.4 \\ 
\midrule
\multicolumn{4}{l}{\textit{\textbf{Open-source Models (7B and 8B)}}} \\
Qwen2.5-VL-7B \cite{bai2025qwen2}    & 83.8 & 35.6 & 89.0 & 75.9 & 4.4 & 79.0 \\
UI-TARS-7B \cite{qin2025ui}    & 63.8 & 2.5 & 80.2 & 11.1 & 3.3 & 8.3 \\ 
OS-Atlas-7B \cite{wu2024atlas}    & 69.6 & 10.0 & 77.0 & 76.8 & 5.5 & 32.1 \\ 
GUI-R1-7B \cite{luo2025gui}     & 70.5 & 7.9 & 84.2 & 45.5 & 5.5 & 8.7 \\ 
UI-Venus-Navi-7B \cite{qin2025ui}   & 89.6 & 53.1 & 92.6 & 91.5 & 45.0 & 81.2 \\ 
AgentCPM-GUI-8B \cite{zhang2025agentcpm}    & 84.9 & 39.0 & 85.8 & 93.8 & 54.9 & 81.2 \\ 
Qwen3-VL-8B \cite{Qwen_Qwen3VL_Collection}     & 91.1 & 59.7 & 92.6 & 82.6 & 69.2 & 94.6 \\
\midrule
\multicolumn{4}{l}{\textit{\textbf{Open-source Models (3B and 4B)}}} \\
Qwen2.5-VL-3B \cite{bai2025qwen2}    & 78.8 & 25.6 & 83.2 & 66.0 & 5.5 & 81.9 \\
UI-R1-3B \cite{lu2025ui}    & 56.4 & 1.0 & 71.2 & 7.6 & 5.5 & 6.5 \\ 
GUI-R1-3B \cite{luo2025gui}    & 45.2 & 0.5  & 57.1 & 8.0 & 2.2 & 4.3 \\ 
Qwen3-VL-4B \cite{Qwen_Qwen3VL_Collection}    & 90.1 & 57.4 & 91.2 & 84.8 & 65.9 & 94.6 \\
\rowcolor{gray!20}
SecAgent-3B (Ours)    &  \textbf{96.4}& \textbf{80.0}& \textbf{96.1} & \textbf{99.1} & \textbf{87.6} & \textbf{99.2}  \\  
\bottomrule
\end{tabular}
\caption{Comparison of various models on the CMGUI benchmark.}
\label{Tab:others_on_benchmark}
\vspace{-12pt}
\end{table*}

The semantic context offers three primary advantages.
First, it only requires single-step historical information instead of the entire one, yielding superior computational efficiency.
Second, the use of natural language to represent historical information provides a more effective mechanism for leveraging the world knowledge embedded within the LLM, which is pre-trained with large linguistic corpora. 
Finally, the semantic context is more informative than sequences of raw, parameterized historical actions (\eg, clicked coordinates), which lack semantic interpretability. 
Without the screenshots on which they are performed, raw actions alone are difficult to provide informative historical information.

\subsection{Training}
To enhance the perception and reasoning capabilities of SecAgent, we employ a two-stage training framework comprising supervised and reinforcement fine-tuning, as illustrated in Fig. \ref{fig:three_stages}.

\mypara{Stage 1: Supervised Fine-Tuning (SFT)} 
We observe that grounding and navigation tasks inherently require different action output formats: grounding tasks require bounding boxes defined by a pair of coordinates, whereas navigation tasks require only a single coordinate representing the click position.
To address this inconsistency, we deviate from prior work \cite{zhang2025agentcpm, zhang2025tongui} that directly utilizes grounding data for SFT training without format alignment.
Instead, we transform the grounding annotations into a navigation-compatible format by computing the center point of each bounding box as the target coordinate for click actions.
This transformation unifies the action space, facilitating better knowledge transfer and generalization between grounding and navigation tasks.
During the SFT stage,
we train on both transformed grounding and navigation data. The model is optimized to output a structured triplet consisting of semantic context, thought, and action, using the next-token prediction loss. 
This stage enhances the agent’s ability to recognize elements and 
enables it to imitate human-like decision-making processes.

\mypara{Stage 2: Reinforcement Fine-Tuning (RFT)}
Following the SFT stage, we further enhance the agent's reasoning capabilities and generalization performance through reinforcement learning.
In the RFT phase, the agent receives a reward for each generated output, and iteratively updates its policy (\ie, model parameters) to maximize the expected reward using the Group Relative Policy Optimization (GRPO) algorithm \cite{shao2024deepseekmath}.
We adopt a rule-based reward comprising format reward and action reward. 
The format reward encourages structural correctness by assigning a value of 0.5 when the agent successfully generates the required triplet format (semantic context, thought, action), and 0 otherwise.
The action reward evaluates functional correctness, providing a reward of 1 for a correct action and 0 otherwise.
\section{Experiments} \label{sec:exp}
\subsection{Experimental Setup}

\mypara{Implementation Details}
We employ Qwen2.5-VL-3B \cite{bai2025qwen2} as the base model to conduct the two-stage training.
We adopt LoRA tuning for the SFT stage, and full-parameter tuning for the RFT stage.
In addition to our CMGUI dataset, we incorporate data from AndroidControl \cite{li2024effects} and GUIOdyssey \cite{lu2024gui} dataset to train our agent.
More training details are provided in the supplementary material.

\mypara{Baselines}
We consider different types of models for comparison: 
(1) Closed-source Models: GPT-4o \cite{hurst2024gpt} and Claude-3-Sonnet \cite{Anthropic_Claude3_2024}; 
(2) General Open-source Models: Qwen2.5-VL \cite{bai2025qwen2} and Qwen3-VL \cite{Qwen_Qwen3VL_Collection}; 
(3) GUI-specific SFT Models: OS-Atlas \cite{wu2024atlas} and UI-TARS \cite{qin2025ui}; 
(4) GUI-specific RFT Models: AgentCPM-GUI \cite{zhang2025agentcpm}, UI-Venus-Navi \cite{qin2025ui}, UI-R1 \cite{lu2025ui}, and GUI-R1 \cite{luo2025gui}.

\mypara{Metrics}
We mainly use two metrics to evaluate the navigation performance of agents: \textbf{Step Accuracy (SA)}, defined as the proportion of correct actions, and \textbf{Task Accuracy (TA)}, defined as the proportion of episodes where the actions are all correct.
\begin{table}[!tbp]
\centering
\footnotesize
\begin{tabular}{lccc}
\toprule
\textbf{Models} & \textbf{AndroidControl} & \textbf{GUIOdyssey}  \\
\midrule
\multicolumn{3}{l}{\textit{\textbf{Closed-source Models}}} \\
GPT-4o \cite{hurst2024gpt} & 20.8$^*$ & 29.3 \\
Claude-3-Sonnet \cite{Anthropic_Claude3_2024} & 12.5$^*$ & 46.5  \\
\midrule
\multicolumn{3}{l}{\textit{\textbf{Open-source Models (7B and 8B)}}} \\
Qwen2.5-VL-7B \cite{bai2025qwen2} & 62.9$^*$ & 59.6  \\
UI-TARS-7B \cite{qin2025ui}   & 74.4$^*$ & 37.5  \\
OS-Atlas-7B \cite{wu2024atlas} & 71.2$^\dagger$ & 65.2  \\
GUI-R1-7B \cite{luo2025gui} & 51.7$^\ddagger$ & 41.9  \\
UI-Venus-Navi-7B \cite{qin2025ui} & \underline{76.1}$^\dagger$ & 71.1  \\
AgentCPM-GUI-8B \cite{zhang2025agentcpm} & 69.2$^*$ & \underline{79.2}  \\
Qwen3-VL-8B \cite{Qwen_Qwen3VL_Collection} & 46.4 & 53.8  \\
\midrule
\multicolumn{3}{l}{\textit{\textbf{Open-source Models (3B and 4B)}}} \\
Qwen2.5-VL-3B \cite{bai2025qwen2} & 60.1 & 49.3  \\
UI-R1-3B  \cite{lu2025ui} & 45.4$^\ddagger$ & 32.7  \\
GUI-R1-3B  \cite{luo2025gui} & 46.6$^\ddagger$ & 34.6  \\
Qwen3-VL-4B \cite{Qwen_Qwen3VL_Collection} & 47.5 & 53.4  \\
\rowcolor{gray!20}
SecAgent-3B (Ours) & \textbf{69.5} & \textbf{74.3} \\
\bottomrule
\end{tabular}
\caption{Performance comparison on two widely used English benchmarks.
We report the step accuracies under high-level instructions. 
Bold text indicates the best score among 3B and 4B models, and underlined text represents the best score among all baseline models.
Superscripts $^*$, $^\dagger$ and $^\ddagger$ indicate results from AgentCPM-GUI \cite{zhang2025agentcpm}, UI-Venus \cite{qin2025ui}, GUI-R1 \cite{luo2025gui}, respectively.}
\label{Tab:comparison_sota}
\vspace{-12pt}
\end{table}
\subsection{Benchmark Results}
\mypara{CMGUI-Bench}
Tab. \ref{Tab:others_on_benchmark} shows the comprehensive evaluation of representative GUI agents on our newly constructed CMGUI-Bench. The results unequivocally demonstrate the superior performance of our proposed SecAgent. SecAgent achieves a remarkable SA of $\mathbf{96.4\%}$ and a dominant TA of $\mathbf{80.0\%}$, significantly outperforming all other closed-source and open-source baselines, regardless of their parameter size. 
Analyzing the action-specific metrics reveals the source of this performance gain: SecAgent achieves high accuracy on the frequently used $\texttt{CLICK}$ action ($\mathbf{96.1\%}$) and the challenging $\texttt{SWIPE}$ action ($\mathbf{87.6\%}$).
This superior performance validates the high quality of the CMGUI dataset and its effectiveness for training agents in the Chinese mobile ecosystem.

\mypara{AndroidControl and GUIOdyssey Benchmark}
To examine the effectiveness of our approach beyond the Chinese ecosystem, we further evaluate it on two widely adopted English mobile navigation benchmarks: AndroidControl \cite{li2024effects} and GUIOdyssey \cite{lu2024gui}.
Note that GUIOdyssey has different evaluation versions. Thus, we use the updated evaluation set and randomly select 100 trajectories from it to evaluate all comparison methods.
As shown in Tab. \ref{Tab:comparison_sota}, SecAgent achieves highly competitive results, securing a SA of $\mathbf{69.5\%}$ on AndroidControl and $\mathbf{74.3\%}$ on GUIOdyssey. Notably, our 3B model outperforms other baseline open-source models in its size category and demonstrates performance comparable to larger models, for instance, exceeding the performance of AgentCPM-GUI-8B on AndroidControl and UI-Venus-Navi-7B on GUIOdyssey.
This highlights that the design of semantic context (even when approximated) and the two-stage training scheme yield efficient and competitive navigation capabilities.

\subsection{Ablation Studies}
\begin{table}[!tbp]
\centering
\footnotesize
\begin{tabular}{l c c  c  c c  }
\toprule
\textbf{Settings} & \textbf{SA} & \textbf{TA} & \textbf{ITC} & \textbf{TTFT} $\downarrow$ & \textbf{TPS} $\uparrow$ \\
\midrule
\textit{N=0} & 85.3 & 36.2 & 1537 & 0.07 & 145 \\
\textit{N=1} & 94.8 & 72.8 & 2239 & 0.11 & 140 \\
\textit{N=2} & 95.1 & 73.8 & 2760 & 0.14 & 134 \\
\textit{N=5} & 95.5 & 74.1 & 3642 & 0.22 & 122 \\
\textit{N=5 w/o SC} & 95.0 & 73.0 & 3507 & 0.22 & 124 \\
\midrule
\textit{N=1 w/o SC} & 90.6& 56.4 & 2171 & 0.11 & 142 \\
\textit{N=1 w/o thought} & 94.0 & 68.9 & 2207 & 0.11 & 141 \\
\bottomrule
\end{tabular}
\caption{Performance and efficiency comparison across different history window sizes ($N$), and the ablation of Semantic Context (SC) and thought.
For efficiency evaluation, we report the Input Token Count (ITC), Time to First Token (TTFT), Tokens per Second (TPS), and training hours, following \cite{lu2024gui}. 
ITC includes both text and vision tokens.
}
\label{Tab:history}
\vspace{-12pt}
\end{table}
\begin{table*}[!thbp]
\centering
\footnotesize
\begin{tabular}{l l C C C C G C G G}
\toprule
\multirow{3}{1cm}{\textbf{Training Data}} & 
\multirow{3}{1cm}{\textbf{Training Method}} & 
\multicolumn{4}{c}{\textbf{In-Domain Apps}} & 
\multicolumn{4}{c}{\textbf{Out-of-Domain Apps}} \\
\cmidrule(lr){3-6} \cmidrule(lr){7-10} 
& & 
\makecell{\textbf{Taobao} \\ SA / TA} & \makecell{\textbf{Kuaishou} \\ SA / TA} & \makecell{\textbf{Fliggy} \\ SA / TA} & 
\makecell{\textbf{JD} \\ SA / TA} &
\makecell{\textbf{Rednote} \\ SA / TA} & \makecell{\textbf{Bilibili} \\ SA / TA} & 
\makecell{\textbf{Baidu Maps} \\ SA / TA} & \makecell{\textbf{Pinduoduo} \\ SA / TA} \\
\midrule
\multirow{2}{*}{Partial} & SFT & 91.2 / 56.9 & 89.1 / 44.8 & 91.9 / 58.3 & 91.9 / 61.1 & 91.7 / 54.5 & 88.4 / 45.8 & 86.6 / 41.7 & 94.4 / 66.7 \\
 & SFT + RFT  & 94.5 / 70.8 & 90.1 / 48.3 & 96.0 / 79.2 & 93.5 / 61.1 & 96.4 / 80.0 & 90.2 / 45.8 & 91.3 / 62.5 & 97.6 / 83.3 \\
\midrule
\multirow{2}{*}{Full} & SFT& 94.5 / 70.8 & 91.2 / 58.6 & 96.1 / 79.2 & 92.9 / 61.1 & 97.3 / 87.3 & 96.3 / 79.2 & \textbf{96.9} / \textbf{79.2} & \textbf{97.7} / 83.3 \\
 & SFT + RFT & \textbf{95.5} / \textbf{73.8} & \textbf{93.2} / \textbf{75.9} & \textbf{96.6} / \textbf{79.2} & \textbf{97.6} / \textbf{83.3} & \textbf{98.4} / \textbf{89.1} & \textbf{97.6} / \textbf{91.7} & 96.6 / \textbf{79.2} & 97.6 / \textbf{88.9} \\
\bottomrule
\end{tabular}
\caption{Performance comparison on in-domain and out-of-domain apps. 
The `Partial' and `Full' training data refer to the entire CMGUI dataset, excluding and including the four out-of-domain apps, respectively.
For presentation conciseness, the results of four representative in-domain apps are presented.}
\label{Tab:in_or_out_domain_v1}
\vspace{-12pt}
\end{table*}
\mypara{Performance and Efficiency Comparison}
Tab. \ref{Tab:history} examines the impact of history window size $N$ (\ie, the number of historical screenshots and actions utilized) and key architectural components (\ie, semantic context and thought) on SecAgent’s performance and efficiency.
Expanding from no history ($N=0$) to a single-step history ($N=1$) yields substantial improvements: SA increases by $9.5\%$ while TA nearly doubles, demonstrating that incorporating history is crucial for successful task completion, especially when some key steps are executed correctly, which can lead to a significant improvement in TA. 
However, further expanding the history window produces diminishing returns.
Increasing $N$ from $1$ to $5$ improves SA and TA by less than $2\%$, while incurring substantial efficiency penalties: ITC grows by $62.7\%$, TTFT doubles, and TPS declines by $12.9\%$. 
These results indicate that $N=1$ achieves a favorable balance between performance and efficiency, and is therefore adopted as the default configuration for SecAgent.
Furthermore, $N=1$ \textit{with SC} achieves performance comparable to $N=5$ \textit{without SC}, while exhibiting significantly higher efficiency, which validates the effectiveness of our SC mechanism.
Ablation results at $N=1$ reveal that both SC and thought mechanisms contribute to performance.
Notably, removing SC causes substantial degradation while yielding negligible efficiency improvements.
This demonstrates that the SC provides critical information concisely without significant computational overhead.

\begin{table}[!tbp]
\setlength{\tabcolsep}{4pt}
\centering
\footnotesize
\begin{tabular}{lccc}
\toprule
\textbf{Training Data} & 
\textbf{Training Method} &
\textbf{SA} & 
\textbf{TA}  \\
 \midrule
Navigation & SFT  & 93.7 & 65.8  \\
Grounding + Navigation  & SFT  & 94.4 & 69.2 \\
Gr2Nav + Navigation  & SFT  & 94.8
& 72.8
\\
Gr2Nav + Navigation & RFT $^*$ & 95.2 & 75.9 \\
Gr2Nav + Navigation & SFT + RFT& \textbf{96.4}& \textbf{80.0}\\
\bottomrule
\end{tabular}
\caption{Comparison of different training data and training methods. `Grounding' denotes the original grounding data, and  `Gr2Nav' denotes the transformed grounding data with navigation-compatible format. $^*$Direct RFT on the base model fails to converge; thus we apply a lightweight SFT with 1k navigation step data before RFT.}
\label{Tab:data_comparison_v1}
\vspace{-12pt}
\end{table}
\mypara{Training Data and Methods}
Tab. \ref{Tab:data_comparison_v1} evaluates the impact of different training data compositions and methodologies. 
First, incorporating grounding data (either original or transformed) improves the performance of SFT-trained agents, demonstrating the quality of our grounding data and its efficacy in enhancing the agent's navigational capabilities. 
Notably, Gr2Nav data provides greater improvement than the original data, validating the effectiveness of our grounding data transformation approach.
Our designed transformation enables consistent alignment between the original grounding-style and navigation-style data.
Second, the comparison between SFT and SFT-initialized RFT approaches demonstrates the potential of RFT, which improves TA by $3.1\%$.
Finally, integrating RFT as a subsequent stage following complete SFT yields the best performance, improving SA by $1.6\%$ and TA by $7.2\%$ over the SFT-only agent. These substantial gains validate the efficacy of our two-stage learning paradigm for achieving proficiency in complex mobile GUI navigation tasks.

\mypara{In-Domain (ID) and Out-of-Domain (OOD) Performance}
Tab. \ref{Tab:in_or_out_domain_v1} reveals the ID and OOD performance under different training data and methods.
First, apps share transferable knowledge. 
When trained on partial data, the model achieves reasonable performance on unseen OOD apps, such as $97.6\%$ SA on Rednote with RFT-augmented training, demonstrating effective zero-shot generalization. 
Moreover, incorporating OOD data improves ID app performance, with Kuaishou TA increasing $13.8\%$ under SFT.
Second, apps possess unique domain-specific knowledge that requires targeted training.
Adding OOD data yields substantial improvements on OOD apps: Bilibili increases TA by $33.4\%$ and BaiduMaps by $37.5\%$ under SFT.
These gains are attributed to the accomplishment of app-specific operations, such as ``sending a danmaku'' in the Bilibili App and ``opening nearby tab'' in the BaiduMaps App, which cannot be inferred from general knowledge alone.
Third, when learning from partial data, the RFT-augmented approach outperforms the SFT-only approach on all OOD apps, demonstrating superior generalization capability.
\section{Conclusion}
In this work, we introduce SecAgent, an efficient mobile GUI agent that addresses two critical challenges: the scarcity of multilingual datasets and inefficient history representation. We contribute a large-scale Chinese mobile GUI dataset and a high-quality navigation benchmark with multi-choice action annotations. Furthermore, we propose a novel semantic context mechanism that significantly enhances computational efficiency while preserving important historical information. Empirical results demonstrate that SecAgent outperforms 3B baselines and achieves performance comparable to 7B-8B models.

{
    \small
    \bibliographystyle{ieeenat_fullname}
    \bibliography{main}
}

\clearpage
\setcounter{page}{1}
\maketitlesupplementary

\section{Details of CMGUI}
\subsection{Action Space}
The action space of CMGUI comprising the following actions: $\texttt{CLICK}$, $\texttt{SWIPE}$, $\texttt{TYPE}$, $\texttt{SYSTEM\_BUTTON}$,
$\texttt{WAIT}$,
and $\texttt{TERMINATE}$.
The description and format of each action are presented in Tab. \ref{Tab:action_space}.

\subsection{App Distribution}
Tab. \ref{tab:app_data} presents the detailed breakdown of trajectory counts for each application in CMGUI.
To ensure comprehensive coverage of real-world mobile usage scenarios, CMGUI comprises 44 widely-used Chinese mobile applications across diverse functional categories, including social media, e-commerce, entertainment, and productivity. 

\subsection{Examples}
\begin{figure*}[!htbp]
\centering
\includegraphics[width=0.98\linewidth]{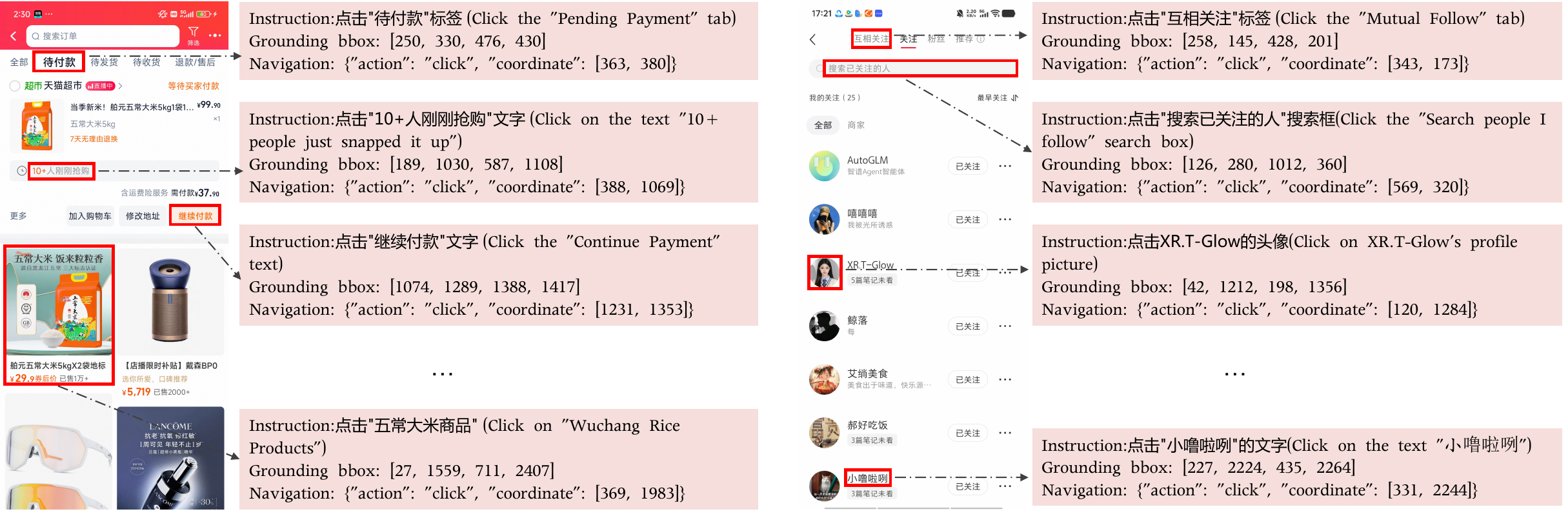}
\vspace{-6pt}
\caption{Examples of grounding data in CMGUI. 
For each screenshot, multiple grounding data can be synthesized and then converted into multiple single-step navigation data.
}
\label{fig:grounding_example}
\vspace{-6pt}
\end{figure*}

Fig. \ref{fig:grounding_example} presents examples of grounding data in CMGUI. 
To improve the utilization of screenshot resources, we generate multiple grounding data from a single screenshot. 
To ensure consistency between grounding and navigation data in output format, we transform the grounding data into a single-step navigation data, with the ground-truth click coordinates corresponding to the bounding box center.

\begin{figure*}[!htbp]
\centering
\includegraphics[width=0.98\linewidth]{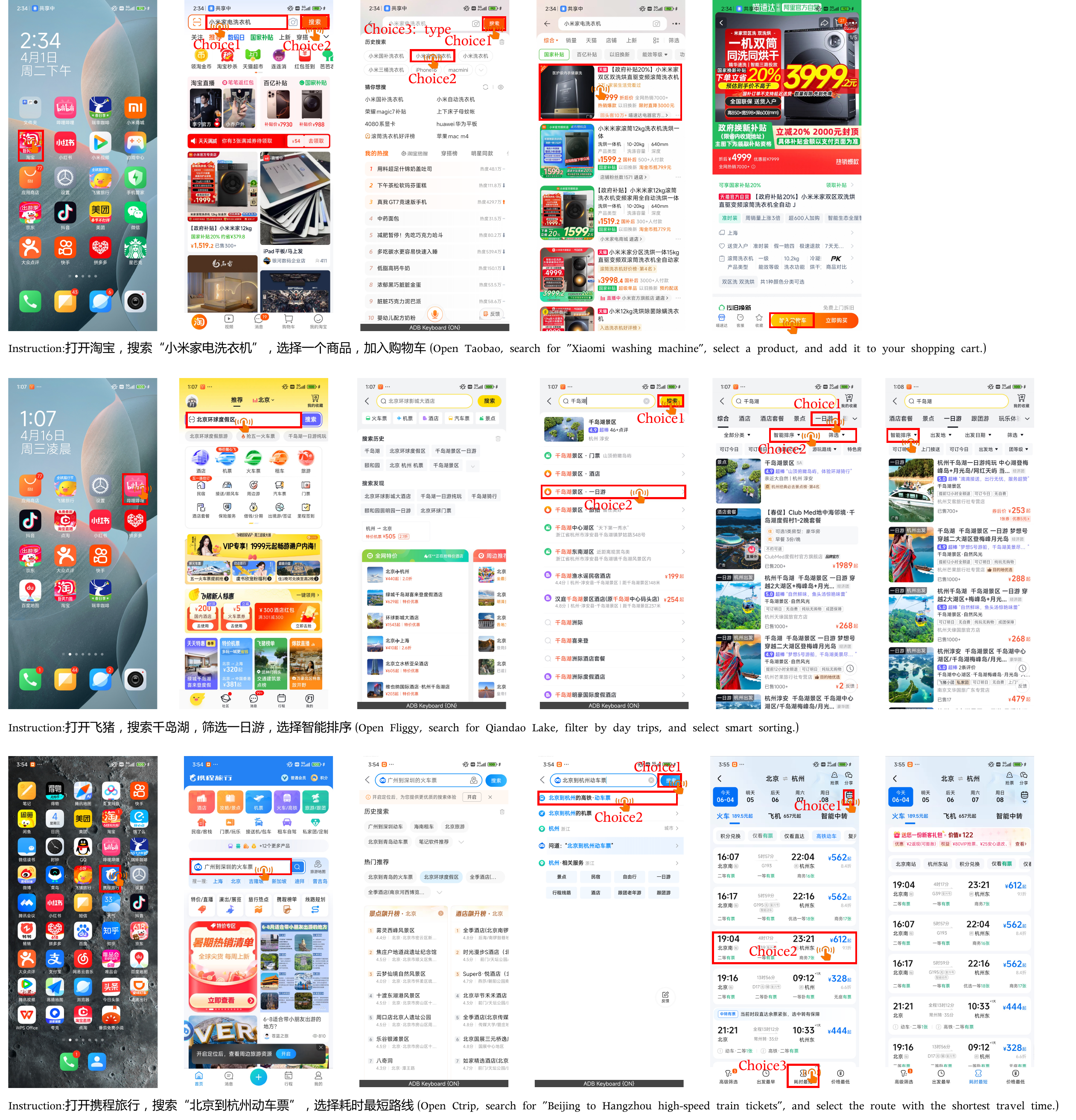}
\vspace{-6pt}
\caption{Example trajectories in CMGUI-Bench. The multi-choice annotations are explicitly drawn on the image.
}
\label{fig:benchmark_example}
\vspace{-6pt}
\end{figure*}

Fig. \ref{fig:benchmark_example} illustrates three evaluation trajectories with multi-choice annotations in CMGUI-Bench.
Given the multi-feasible nature of GUI manipulations, a single step may have multiple correct actions that lead to the same goal. 
To address this, we manually examined each step and annotated all valid actions.

\section{Experiment Details}

\subsection{Prompts}
\begin{figure*}[!htbp]
\centering
\includegraphics[width=0.98\linewidth]{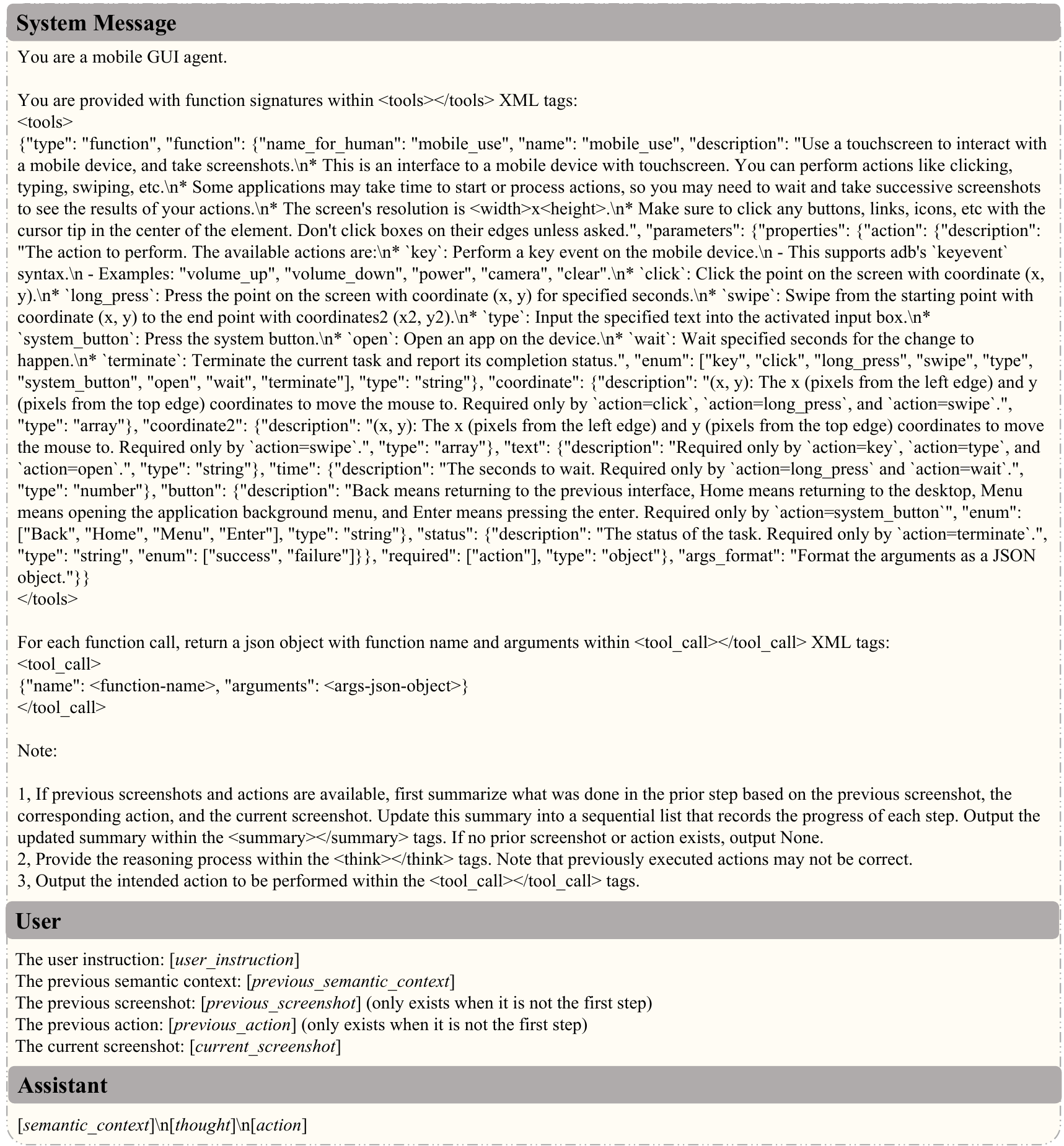}
\vspace{-6pt}
\caption{The prompt used by SecAgent. 
We note that the Chinese text has been translated into
English for presentation purposes.
}
\label{fig:training_prompt}
\vspace{-6pt}
\end{figure*}

Fig. \ref{fig:training_prompt} shows the prompt used by SecAgent.
The prompt includes the action space specification, user instruction, previous semantic context, current screenshot, and previous action as inputs. The agent is required to output the updated semantic context, its thought, and the action to take.
For the baseline models, we strictly follow their respective official guidelines and prompt templates when available.

\begin{figure*}[!htbp]
\centering
\includegraphics[width=0.94\linewidth]{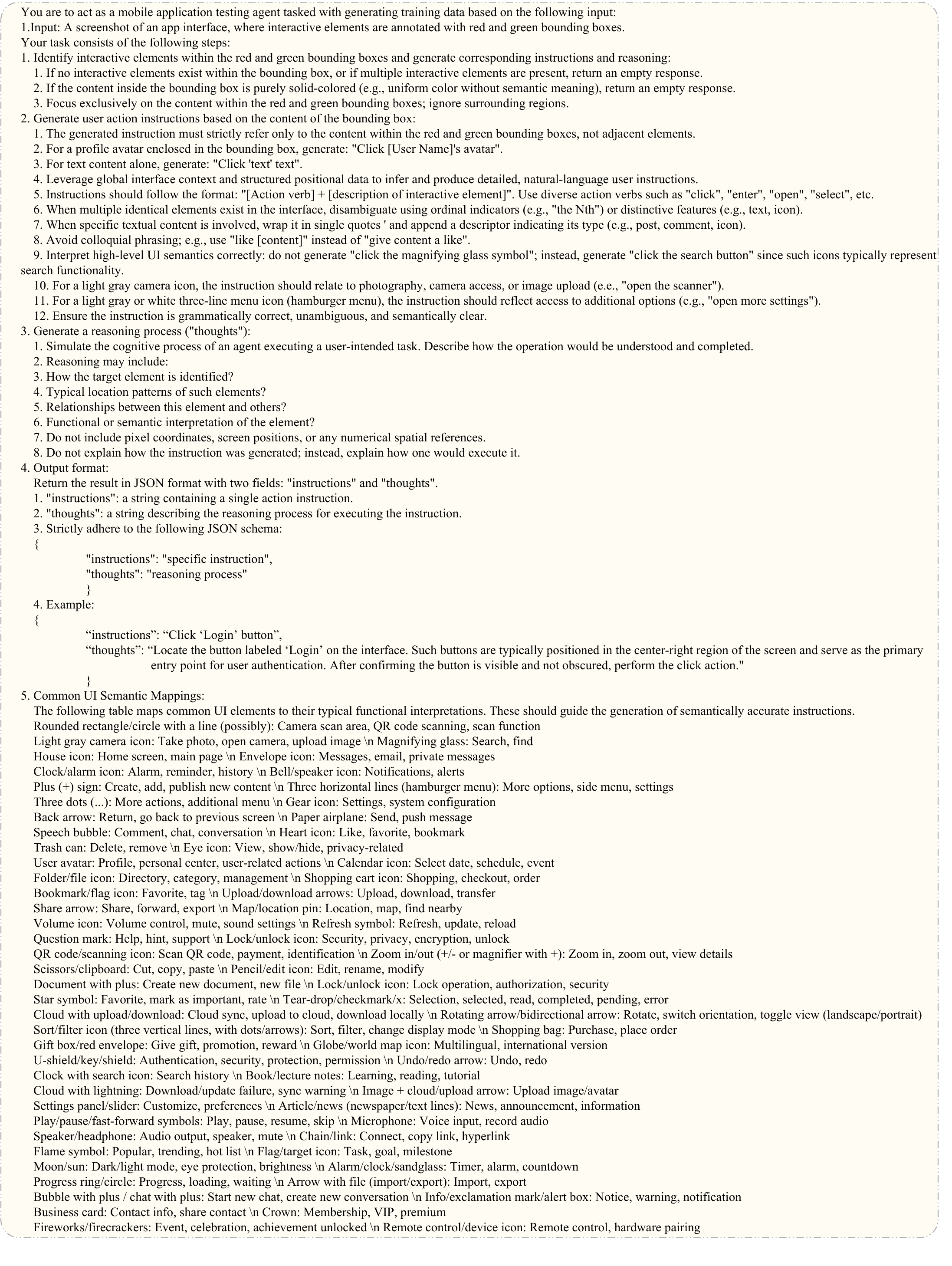}
\vspace{-20pt}
\caption{The prompt used to generate grounding instructions.
We note that the Chinese text has been translated into
English for presentation purposes.
}
\label{fig:grounding_prompt}
\vspace{-6pt}
\end{figure*}

The prompt used to generate grounding instructions is shown in Fig. \ref{fig:grounding_prompt}. 
In the prompt, we specify the instruction format, require the agent to provide explanations for its generated instructions, and provide descriptions of common UI elements and their meanings.

\subsection{RFT Algorithm}
Our RFT algorithm is built on the GRPO algorithm \cite{shao2024deepseekmath},
which stabilizes and accelerates training by estimating the advantage of actions using group-normalized rewards without acquiring a reward model or critic model.
We adopt two modifications to the GRPO algorithm:  the token-level loss and different clipping ratios.
Given a query $q$, the current policy $\pi_{\theta_{\mathrm{old}}}$ samples $G$ responses $\{o_i\}_{i=1}^{G}$. Each response is assigned a scalar reward $\{R_i\}_{i=1}^{G}$. Rewards are normalised within the group to produce variance-reduced advantages:
\vspace{-4pt}
\begin{equation}
\vspace{-4pt}
    \hat{A}_{i, t} = \frac{R_i - \mathrm{mean}(\{R_i\}_{i=1}^{G})}{\mathrm{std}(\{R_i\}_{i=1}^{G})}
    \label{eq:advantage}
\end{equation}
The policy is then updated using a clipped objective with KL divergence penalty:
\vspace{-4pt}
\begin{equation}
\vspace{-4pt}
\begin{split}
J_{\mathrm{GRPO}}(\theta) =& \frac{1}{\sum_{i=1}^{G}|o_i|} \sum_{i=1}^{G} \sum_{t=1}^{|o_i|} \Big\{\min  \left( r_{i,t}(\theta) \hat{A}_{i,t},  \right.  \\
& \left. \mathrm{clip}\left( r_{i,t}(\theta), 1 - \epsilon_{\text{low}}, 1 + \epsilon_{\text{hight}} \right) \hat{A}_{i,t} \right) \\
& - \beta \mathbb{D}_{KL}\left[ \pi_\theta \| \pi_{\mathrm{ref}} \right] \Big\}
\end{split}
\end{equation}
where $\epsilon_{\text{low}}$ and $\epsilon_{\text{hight}}$ control the clipping range, 
$\beta$ constrains policy divergence from the reference model $\pi_{\mathrm{ref}}$, and
$r_{i,t}(\theta)$ is the probability ratio, calculated as:
\vspace{-4pt}
\begin{equation}
\vspace{-4pt}
r_{i,t}(\theta) = \frac{\pi_\theta(o_{i,t} \mid q, o_{i,<t})}{\pi_{\theta_{\mathrm{old}}}(o_{i,t} \mid q, o_{i,<t})}
\end{equation}

\subsection{Training Parameters}
The hyperparameters used for training SecAgent during SFT and RFT are shown in Tab. \ref{tab:parameters}.
We use Qwen2.5-VL-3B as our base model, and set \texttt{min\_pixels} and \texttt{max\_pixels} as $200,704$ and $501,760$, respectively.
\begin{table}[hbtp]
\vspace{-5pt}
\centering
\footnotesize
\begin{tabular}{ll}
\toprule
\textbf{Hyperparameter} & \textbf{Value} \\
\midrule
\multicolumn{2}{l}{\textit{\textbf{Supervised Fine-tuning}}} \\
Epoch & 6 \\
Batch Size & 2 \\
Learning Rate & 2e-5 \\
LoRA Rank & 8 \\
LoRA Alpha & 16 \\
\midrule
\multicolumn{2}{l}{\textit{\textbf{Reinforcement Fine-tuning}}} \\
Epoch & 2 \\
Batch Size & 8 \\
Learning Rate & 1e-6 \\
KL Penalty & 0.04 \\
Number of Rollouts & 16 \\
Clip Range $\epsilon_{\text{low}}, \epsilon_{\text{high}}$ & 0.2, 0.28 \\
\bottomrule
\end{tabular}
\caption{The hyperparameters used for training SecAgent.}
\label{tab:parameters}
\vspace{-30pt}
\end{table}

\begin{table*}[!htbp]
\centering
\footnotesize
\begin{tabular}{lp{4cm}p{8cm}}
\toprule
\textbf{Action} & 
\textbf{Description} &
\textbf{Format} \\
 \midrule
\texttt{CLICK}& Click the point on the screen with coordinate (x, y).& \texttt{\{"name": "mobile\_use", "arguments": \{"action": "click", "coordinate": [x, y]\}\}}\\
\texttt{SWIPE}& Swipe from the starting point with coordinate (x, y) to the end point with coordinates2 (x2, y2).& \texttt{\{"name": "mobile\_use", "arguments": \{"action": "swipe", "coordinate": [x, y], "coordinate2": [x2, y2]\}\}} \\
\texttt{TYPE}& Input the specified text into the activated input box. & \texttt{\{"name": "mobile\_use", "arguments": \{"action": "type", "text": "specified\_text"\}\}} \\
\texttt{SYSTEM\_BUTTON}& Press the specified  system button (\eg, ``Back'', ``Home'', ``Menu'', ``Enter''). & \texttt{\{"name": "mobile\_use", "arguments": \{"action": "system\_button", "button": "specified\_button"\}\}} \\
\texttt{WAIT}& Wait specified seconds for the change to happen. & \texttt{\{"name": "mobile\_use", "arguments": \{"action": "wait", "time": "specified\_seconds"\}\}} \\
 \texttt{TERMINATE}& Terminate the current task and report its completion status (``success'' or ``failure''). & \texttt{\{"name": "mobile\_use", "arguments": \{"action": "terminate", "status": "success"\}\}} \\
\bottomrule
\end{tabular}
\caption{The action space of CMGUI.}
\label{Tab:action_space}
\vspace{-12pt}
\end{table*}

\begin{table*}[!htbp]
  \centering
  \footnotesize
  \begin{tabular}{clcclcclcclc}
    \toprule
     \#ID &\textbf{App Name} & \textbf{\#Traj.} & \#ID &\textbf{App Name} & \textbf{
     \#Traj.}  &\#ID &\textbf{App Name} & \textbf{\#Traj.} &\#ID &\textbf{App Name} & \textbf{\#Traj.}\\
    \midrule
1   &Taobao & 4,722 
& 2    &Kuaishou & 1,829
& 3    &Rednote & 1,341 
& 4  & Baidu Maps & 1,213  \\
5    &Fliggy & 1,184
&6   &Bilibili & 1,157 
&7  &WeChat & 1145
&8    &JD & 1,023  \\
9    &Pinduoduo & 861 
&10 &Weather & 817 
&11 &Meituan & 777 
&12  &Douyin & 757 \\
13   &Calendar & 741
&14    &Poizon & 714
&15     &Aliyun Drive & 693 
&16     &Amap & 686 \\
17     &Ele.me & 640
&18   &Quark Browser & 589 
&19     &Toutiao & 580
&20 &Tencent Maps & 576 \\
21   &Calculator & 541 
&22    &Baidu & 519 
&23 &Tencent Meeting & 499 
& 24    &Browser & 496 \\
25    &Tomato Novel & 492 
& 26    &Didi Chuxing & 447
& 27    &Notes & 432 
&28    &WeCom & 429 \\
29 &Message & 426 
&30    &Contacts & 401 
&31 &DingTalk & 393
&32    &Feishu & 380 \\
33    &Baidu Netdisk & 355
&34     &Cainiao &  329
&35  &Ctrip & 264
&36    &WPS Office & 163 \\
37 &QQ & 152 
&38 &WeChat Reading & 151
&39 &Weibo & 149 
&40 &Zhihu & 130 \\
41   &Xianyu & 129   
& 42    &Alipay &  120 
&43    &Zhuanzhuan & 117 
& 44    &Luckin Coffee& 48 \\
        \bottomrule
  \end{tabular}
\caption{Our constructed CMGUI consists of 44 mainstream Chinese apps. The number of trajectories for each app is shown in the table.}
\label{tab:app_data}
\end{table*}

\end{document}